\documentclass[10pt,twocolumn]{article}

\usepackage[top=0.75in, bottom=1in, left=0.65in, right=0.65in, columnsep=0.2in]{geometry}
\usepackage{amsmath,amssymb}
\usepackage{graphicx}
\usepackage{authblk}
\usepackage{multicol}
\usepackage[colorlinks=true,linkcolor=blue,citecolor=blue,urlcolor=blue]{hyperref}
\usepackage[numbers]{natbib}

\usepackage{amsmath}
\usepackage{amssymb,amsfonts}
\usepackage{graphicx}
\usepackage{textcomp}
\usepackage{comment}
\usepackage{algorithm}
\usepackage{array}
\usepackage{subcaption}
\usepackage{pgffor}

\usepackage{siunitx}
\usepackage{subcaption}
\usepackage{mathtools}
\usepackage{algorithm}
\usepackage{algorithmic}
\usepackage{booktabs}
\usepackage{todonotes}

\newcommand{\Langevin}{\mathcal{L}}
\newcommand{\regularizer}{\mathcal{R}}
\newcommand{\interpolator}{\mathcal{I}}

\newcommand{\hsat}{H_{\mathrm{sat}}}
\newcommand{\msat}{M_{\mathrm{sat}}}
\newcommand{\Hxt}{H(x,t)}
\newcommand{\Mxt}{M(x,t)}
\newcommand{\Hxtn}{\frac{H(x,t)}{\lVert H(x,t)\rVert}}
\newcommand{\Hhsat}{\frac{\lVert H(x,t)\rVert}{\hsat}}

\newcommand{\norm}[1]{\left\lVert #1 \right\rVert}

\DeclareMathOperator{\diag}{diag}
\DeclareMathOperator{\trace}{tr}

\title{Fast Trajectory-Independent Model-Based Reconstruction Algorithm for Multi-Dimensional Magnetic Particle Imaging}

\author[1,2]{Vladyslav Gapyak \thanks{Emails: vladyslav.gapyak@h-da.de, thomas.maerz@h-da.de, andreas.weinmann@h-da.de}}
\author[1,2]{Thomas März}
\author[1,2]{Andreas Weinmann}
\affil[1]{Algorithms for Computer Vision, Imaging and Data Analysis Lab at Darmstadt University of Applied Sciences, Sch\"{o}fferstr. 3, 64295, Darmastdt, Germany}
\affil[2]{Data Science Institute, European University of Technology, European Union}

\date{\today}

\begin{document}
	\maketitle
	
	\begin{abstract}
			Magnetic Particle Imaging (MPI) is a promising tomographic technique for visualizing the spatio-temporal distribution of superparamagnetic nanoparticles, with applications ranging from cancer detection to real-time cardiovascular monitoring. Traditional MPI reconstruction relies on either time-consuming calibration (measured system matrix) or model-based simulation of the forward operator. Recent developments have shown the applicability of Chebyshev polynomials to multi-dimensional Lissajous Field-Free Point (FFP) scans. This method is bound to the particular choice of sinusoidal scanning trajectories. In this paper, we present the first reconstruction on real 2D MPI data with a trajectory-independent model-based MPI reconstruction algorithm. We further develop the zero-shot Plug-and-Play (PnP) algorithm of the authors -- with automatic noise level estimation -- to address the present deconvolution problem, leveraging a state-of-the-art denoiser trained on natural images without retraining on MPI-specific data. We evaluate our method on the publicly available 2D FFP MPI dataset “MPIdata: Equilibrium Model with Anisotropy,” featuring scans of six phantoms acquired using a Bruker preclinical scanner. Moreover, we show reconstruction performed on custom data on a 2D scanner with additional high-frequency excitation field and partial data. Our results demonstrate strong reconstruction capabilities across different scanning scenarios—setting a precedent for general-purpose, flexible model-based MPI reconstruction.
	\end{abstract}
	\vspace{1em}
	

\section{Introduction}\label{sec:introduction}

Magnetic Particle Imaging (MPI) is an imaging modality introduced by Gleich and Weizenecker \cite{gleich2005original} in 2005, characterized by its rapid acquisition time, high sensitivity and absence of ionizing radiation. It allows to map the spatio-temporal concentration distribution of superparamagnetic nanoparticles like iron oxide nanoparticles (SPIOs) injected into the target specimen. In 2009 Weizenecker et al. \cite{Weizenecker_etal2009} produced a video (3D+time) of the distribution of particles in the beating heart of a mouse; this result has been a milestone towards the application of MPI to real-time \emph{in vivo} imaging. Since then, a variety of different applications have been shown to benefit from MPI; among these applications we mention multimodal imaging~\cite{aramiMultimodalImagingMPI}, detection of as few as 250 cancer cells~\cite{Song2018} and cancer imaging~\cite{aramiMultimodalImagingMPI,Du2019,Tay2021,Yu2017}, (stem) cell tracing~\cite{connell2015advancedcellTherapies,GoodwillConolly2011,JUNG2018139,Lemaster2018,Tomitaka2015Lactoferrin}, inflammation tracing and lung perfusion imaging~\cite{Zhou_2017}, drug delivery and monitoring~\cite{Zhu2019}, cardiovascular~\cite{Bakenecker2018MPIvascular,Tong2021,Vaalma2017} and blood flow~\cite{Franke2020BloodFlow} imaging, tracking of medical instruments~\cite{haegele2012instrumentvisualization}, brain injury detection~\cite{Orendorff2017,Graeser2019humanbrain}. To obtain the mapping of the particles, dynamic magnetic fields are applied to the specimen inside the scanner and the change in magnetic moments of the particles induces a voltage captured by receiving coils. This induced voltage constitutes the MPI signal from which the particles distribution has to be reconstructed.  As of today, the main reconstruction approaches are either measure-based approaches and model-based approaches. Measure-based approaches usually work by considering a grid discretization of the Field of View (FoV) and by scanning a probe containing a known concentration of particles in each voxel of the grid. Thus, the system's response to delta impulses is collected in a system matrix that describes the forward operator underlying the scanning procedure. Regularized inversion of the system matrix is the the most popular reconstruction approach and remains an active area of research \cite{Knopp_etal2010ec,storath2016edge,gapyak2025ell1pnp,dittmer2020deep,askin2022pnp,gungor2023deqmpi}. 
To avoid time-consuming calibration procedures, model-based approaches have been studied in MPI. Such approaches can be subdivided into two main branches: (i) in the first, the system matrix is simulated and the reconstruction is achieved with the techniques developed for the measured system-matrix-based approaches \cite{maass2024equilibriumanysotropy}; (ii) in the second, the reconstruction is performed by describing the full forward operator with a mathematical model. The models employed and proposed range from the Equilibrium (Langevin) Model to the more complex Equilibrium Models with Anisotropy \cite{maass2024equilibriumanysotropy} or with a modified Jiles–Atherton model \cite{li2023jiles-arthenton}. 

Within model-based approaches, the X-Space formulation by Goodwill and Conolly \cite{GoodwillConolly2010,GoodwillConolly2011} operates in the time domain, enabling reconstruction for 1D scanning sequences. Their applicability to multi-dimensional scanning sequences  -- like 2D and 3D Lissajous trajectories -- has been demonstrated when one of the frequencies is much higher than the others, i.e., in a setup that approximates 1D scans. Consequently, the applicability of the X-Space method on general Lissajous (and non-Lissajous) scanning trajectories on real data remains open. Recent advances by Sanders et al. \cite{sanders2025physicsbased} have shown that it is possible to obtain high-quality reconstructions in Field-Free Line (FFL) MPI by implementing a computational physics model. In this model, the scanning forward operator is formed by the multiplication of a series of linear operators that can be calibrated with reduced effort, when compared to the measured system-matrix approach. While the reconstructions are of high quality, the applicability of the method in \cite{sanders2025physicsbased} to multi-dimensional Field-Free Point (FFP) MPI is only shown in simulated scenarios.

The model-based reconstruction problem can also be analyzed in frequency domain. In the Fourier domain, Chebyshev polynomials have been used to directly reconstruct concentrations from 1D scanning trajectories \cite{erb2018mathematical}. Recently, direct reconstruction with Chebyshev polynomials has been achieved in multi-dimensional FFP MPI on 2D Lissajous trajectories on real data \cite{droigk2022multidimcheb}. The decomposition in Chebyshev polynomials arise due to the sinusoidal nature of Lissajous trajectories. Its applicability to reconstruction with other and more general scanning trajectories remains open.

In this paper we show the first model-based reconstructions on real data obtained with a trajectory-independent method, suggesting its potential applicability to more general MPI scanning scenarios.

\subsection{Contributions}

We show for the first time that model-based reconstructions in a 2D FFP setting on real MPI data are possible with an algorithm independent from the chosen scanning trajectory. The theoretical background of the proposed method stems from the authors' paper in \cite{marz2016model}. The method proceeds in two stages: (i) a Core Stage, where the so-called MPI Core Operator is reconstructed starting from the data and (ii) a deconvolution stage. For the Core Stage, we use a variational formulation \cite{gapyak2022mdpi}.

We leverage recent developments in Machine-Learning-based methods for the deconvolution problem, which is fundamental for model-based MPI \cite{maass2024equilibriumanysotropy,sanders2025physicsbased}. Indeed, we propose to use a Plug-and-Play algorithm for the deconvolution stage. Plug-and-Play approaches \cite{venkatakrishnan2013pnp} are approaches that allow to solve a regularized inverse problem iteratively by alternating two steps: first, solve a Tikhonov type problem, and second, apply a Gaussian denoiser. This approach has been used in measured system-matrix approaches (cf. \cite{gapyak2025ell1pnp}) in combination with the \emph{deep denoiser prior} \cite{Zhang2022pnp} in a zero-shot fashion. We here further develop this approach, and introduce a ZeroShot-PnP deconvolution algorithm with automatic noise estimation that leverages the image denoising capabilities of denoisers trained on large datasets of natural images. In this way, we avoid the training of the Gaussian denoiser on the MPI specific domain.

We show reconstruction results on the 2D FFP ``MPIdata: Equilibrium Model with Anisotropy" dataset \cite{knopp2024equilibriumdata} published with \cite{maass2024equilibriumanysotropy}. This dataset consists of six phantoms scanned with the preclinical MPI scanner from Bruker (Ettlingen, Germany). The scanner employs a Lissajous trajectory for the scan. We compare our reconstruction results with reconstructions obtained with state-of-the-art methods. Additionally, we show reconstructions on custom data that employ a non-Lissajous trajectory and with scan data coming from only one channel, resulting in partial data. We show that also in this case the reconstruction is possible and of competitive quality.

\section{Methods}\label{sec:theory}

In MPI, the signal $s(t)$ acquired during a scan and background corrected can be modeled as\cite{kluth2018mathematical}
\begin{equation}\label{eq:signal}
	s(t) = -\mu_0 \frac{d}{dt}\left [\int_{\mathbb{R}^n}\rho (x) R(x) \Mxt\, dx\right ] * a(t)
\end{equation}
where $\rho$ is the target particle concentration, $\mu_0$ is the magnetic permeability in vacuum, $\Hxt$ is the magnetic field applied, $\Mxt$ is the magnetization response of the particles, $R$ is the matrix characterizing the sensitivity profile of the receiving coils, and $a$ is a periodic vector kernel of the analog filter in the signal acquisition chain convolved componentwise. Using the Langevin model of paramagnetism (equilibrium isotropic approximation) the magnetization response can be written as
\begin{equation}\label{eq:momentum}
	M(x,t) = m\rho (x) \Langevin \left (\Hhsat\right )\Hxtn
\end{equation}
where $\Langevin = \cosh (z)- \frac{1}{z}$ is the Langevin function and $\hsat$ is the saturation parameter that encapsulates both the size of the particles as well as other physical parameters
\begin{equation}\label{eq:hsat}
	\hsat = \frac{\kappa_b T}{\mu_0 \msat \frac{\pi}{6}d^3} .
\end{equation}
In \eqref{eq:hsat} $\kappa_b$ is Boltzmann's constant, $T$ is the temperature of the particles, $\msat$ is the magnetic saturation and $d$ is the diameter of the particles. 

Before proceeding with the theoretical background, we introduce the MPI Kernel and the MPI Core Operator, which constitute the fundamental mathematical objects behind the model-based MPI reconstruction here presented.

We define the MPI convolution kernel as the function $K\colon \mathbb{R}^n\mapsto\mathbb{R}^{n\times n}$ defined by the position
\begin{equation}\label{eq:kernel:matrix}
	K(y) = \Langevin '(\norm{y})\frac{yy^T}{\norm{y}^2} + \frac{\Langevin (\norm{y})}{\norm{y}}\left [ \mathbb{I}- \frac{yy^T}{\norm{y}^2} \right ],
\end{equation}
and its resolution-rescaled version as $K_h (y) = \frac{1}{h}K(\frac{y}{h})$. Given a particles concentration $\rho$, the MPI Core Operator is defined by its convolution with the MPI Kernel
\begin{equation}\label{eq:coreoperator}
	A_h [\rho ] (x) = (K_h * \rho ) (x) = \int_{\mathbb{R}^3} \rho (y) K_h (x-y)\, dy .
\end{equation}
These operators are fundamental in the description of the signal as they arise from the time derivative of the magnetization, i.e., 
\begin{align*}
	\frac{d}{dt} & \int_{\mathbb{R}^n}\rho (x) R(x) \Mxt\, dx  = \int_{\mathbb{R}^n}\rho (x) R(x)\frac{\partial}{\partial t}\Mxt\, dx \\
	= & \int_{\mathbb{R}^n}\rho (x) R(x) K_{\hsat}\left ( \Hxt\right )\dot{H}(x,t)\, dx .
\end{align*}
In the Field-Free Point (FFP) configuration, the dynamic magnetic field applied consists fo the superposition of a static (selection) field $H_s (x) = Gx$ with gradient $G$, and a dynamic (drive) field $H_d (t)$. The total magnetic field applied can be written simply as $\Hxt = H_s (x) + H_d (t) =  G(x-r(t))$ where $r(t) = -G^{-1}H_d (t)$ is the scanning trajectory and the trajectory of the FFP. 
Additionally, with the standard choice of a diagonal gradient , e.g., $G = \diag (-1, -1,2)\si{\tesla\per\mu_0\meter}$ and a constant sensitivity profile $R(x)=R$, the signal can be modeled as a double convolution
\begin{equation}\label{eq:signal:noA}
	s(t) = -\mu_0 m R  \left [\left ( \rho *K_{\hsat}\right ) (r(t )) \, v(t) \right ]* a (t) .
\end{equation}

Consequently, the relationship in \eqref{eq:signal:noA} can be rewritten as
\begin{equation}\label{eq:core:formula:conva}
	s(t) \propto  \left [A_{\hsat} [\rho ] \left ( r(t) \right ) v(t)\right ] * a(t)
\end{equation}
where the proportionality constant is  $ -\mu_0 m R$. 

The time-convolution with the kernels $a$ is usually taken care of in a pre-processing step. The Fourier transformed transfer function $\hat{a}$ is collected and used for division of the signal in Fourier domain $\hat{s}/\hat{a}$. This corresponds to the deconvolution in time domain. As a result, we will consider the transfer-function-corrected signal 
\begin{equation}\label{eq:core:formula}
	s(t) \propto  A_{\hsat} [\rho ] \left ( r(t) \right ) v(t)
\end{equation}
for all methods described in the reconstruction algorithm.

\subsection{MPI Core Stage}

The Core Stage is the first reconstruction step of our algorithm and stems from the relationship in \eqref{eq:core:formula}. In particular, in the Core Stage we reconstruct the matrix-valued MPI Core Operator $A_{\hsat}\in\mathbb{R}^{n\times n}$ from the data $(s_k , r_k , v_k )$ in a variational fashion \cite{gapyak2022mdpi}, i.e., by minimizing the functional

\begin{align}\label{eq:first:step}
	A = \arg\min\limits_{\hat{A}}\left\lbrace 
	\frac{1}{L} \sum\limits_{k=1}^L \left\lVert s_k - \interpolator \bigl [\hat{A}\bigr ](r_k) v_k \right\rVert^2 
	+ \gamma\regularizer_C [\hat{A}]
	\right\rbrace ,
\end{align}
where $\regularizer_C$ is a regularization term, $\interpolator$ is a chosen interpolation scheme and $\gamma >0$ is a parameter controlling the strength of the regularization. In this paper, we consider the cosine interpolation scheme. The MPI Core Operator is a smooth and analytic function; consequently, we enforce smoothness by considering a second-order regularization term, i.e., the $L^2$-norm of the Laplacian. Numerically, the minimization in \eqref{eq:first:step} is achieved by considering the Euler-Lagrange equations and using the Conjugated Gradient (CG) algorithm as an iterative solver. The tolerance for the CG method has been set to $10^{-3}$ on the relative norm of the residual and a maximum of 10,000 iterations. The regularization parameter $\gamma$ is set to $10^{-7}$ for all phantoms and experiments. The parameter has been selected by visual inspection of the results.

\subsection{Deconvolution Stage}

In \cite{marz2016model} it has been shown that the trace of the MPI Core Operator contains the information necessary to retrieve the particles concentration, i.e.,
\begin{equation}\label{eq:convolution:trace}
	\trace A_{h}[\rho ] (x) = (\kappa_h * \rho )(x)
\end{equation}
where $\kappa_h$ is the trace of the kernel in \eqref{eq:kernel:matrix},
\begin{equation}\label{eq:kernel:trace}
	\kappa_h (y) = \Langevin '(\norm{y}) + \frac{\Langevin (\norm{y})}{\norm{y}}(n-1) .
\end{equation}
Here $n$ is the dimensionality of the scan, $h$ is some resolution parameter (usually $h=\hsat$) and $\kappa_h$ is a positive definite functional (cf. \cite{gapyak2023ffl3d}).
It follows, that the final reconstruction of the particles concentration can be obtained via a scalar deconvolution with $\kappa_h$. We perform deconvolution using a Plug-and-Play (PnP) approach. This is inspired by variational regularization solving the minimization problem
\begin{equation}\label{eq:dec:min}
	\rho = \arg\min_{\hat{\rho}}\lVert\kappa_h * \hat{\rho} - u\rVert_2^2 + \lambda \regularizer_D [\hat{\rho}],
\end{equation}
for some regularization prior $\regularizer_D$ and regularization parameter $\lambda> 0$. Half Quadratic Splitting (HQS) of the regularized minimization problem in \eqref{eq:dec:min} considers the equivalent constrained minimization problem
\begin{equation}\label{eq:dec:min:2}
	\rho = \arg\min_{\rho_1 \, , \rho_2}\lVert\kappa_h * \rho_1 - u\rVert_2^2 + \lambda \regularizer_D [\rho_2]\quad\text{s.t. } \rho_1 = \rho_2\, ,
\end{equation}
and alternatingly minimizes the Lagrangian in the two variables $\rho_1$ and $\rho_2$:
\begin{align}
	\rho_1^{k+1} & = \arg\min_{\rho_1}\lVert \kappa_h * \rho_1 -u\rVert_2^2 +  \nu_k\left\lVert \rho_1-\rho_2^k \right\rVert_2^2 \label{eq:dec:sub:tik0}\\
	\rho_2^{k+1} & = \lambda \regularizer_D [\hat{\rho}_2]+\nu_k\left\lVert \rho_1^{k+1}-\rho_2 \right\rVert_2^2 . \label{eq:dec:sub:den:gauss}
\end{align}
The term in \eqref{eq:dec:sub:den:gauss} can be interpreted as a Gaussian denoising problem of the variable $\rho_1^{k+1}$ with noise level $\sqrt{\lambda /\nu_k}$. We adapt the algorithm in \cite{gapyak2025ell1pnp} and substitute \eqref{eq:dec:sub:den:gauss} with the \emph{deep denoiser prior}\cite{Zhang2022pnp}, a pre-trained Gaussian denoiser that can take into account the constant $\sqrt{\lambda /\nu_k}$ as noise level map.
More specifically, the minimization problem in \eqref{eq:dec:min} is solved by alternating between a simpler Tikhonov data-fidelity step and a Gaussian denoising step
\begin{align}
	\rho_1^{k+1} & = \arg\min_{\rho_1}\lVert u-C_h \rho_1\rVert_2^2 +  \nu_k\left\lVert \rho_1-\rho_2^k \right\rVert_2^2 \label{eq:dec:sub:tik}\\
	\rho_2^{k+1} & = \mathrm{Denoiser}\left (\tilde{\rho}_1^{k+1}\, , \sqrt{\lambda /\nu_k}\right ) , \label{eq:dec:sub:den}
\end{align}
where $C_h$ is the discretized version of the convolution operator with kernel $\kappa_h$ and $\tilde{\cdot}$ represent the reshaping operator, converting vectors in 2D-arrays.
The noise level $\sigma_{k+1}=\sqrt{\lambda /\nu_k}$ lays in inverse relation with the Tikhonov parameter $\nu_k$. This property allows to only treat $\nu_0$ as a (hyper)parameter, as the $\sigma_{k+1}$ can be estimated at each iteration from the iterate $\rho_1^{k+1}$ \cite{gapyak2025ell1pnp}. We have chosen to estimate $\sigma_{k+1}$ with the root of the variance 
\begin{equation}
	\sigma_{k+1} = \sqrt{\mathrm{Var}(\rho_1^{k+1})}
\end{equation}
of the data-fidelity reconstruction $\rho_1^{k+1}$. Although simple, this choice has proven itself reliable and yielded good results in the reconstruction.

Upon discretization, the deconvolution problem \eqref{eq:dec:min} is formally identical to a regularized inversion of the Toepliz matrix representing the convolution with the kernel $\kappa_h *\rho$. Because the discretized kernel $\kappa_h$ is of the same size of the variable $\rho$ we can write the problem using a sparse format for the matrix and use the FFT for the convolution by direct direct multiplication in Fourier domain. A pseudo-algorithm for the ZeroShot-PnP deconvolution can be seen in Algorithm \ref{alg:pnp}. 

Concerning the numerical implementation, to solve the minimization problem in \eqref{eq:dec:sub:tik} we use the CG method on the Euler-Lagrange equations with a tolerance of $10^{-3}$ on the relative norm of the residual and a maximum number of 10,000 iterations. The starting $\nu_0$ parameter is set to $10^{-5}$ for all phantoms but the snail phantom for which we used $10^{-6}$. The parameter has been selected by visual inspection of the results.

\begin{algorithm}
	\caption{\small The proposed deconvolution with the ZeroShot-PnP algorithm.}\label{alg:pnp}
	\textbf{Input}: trace $u=\trace A$ and  $\kappa_h$, parameters, $n_{\mathrm{it}}$, $\nu_0$.\\
	\textbf{Output}: reconstructed concentration $\tilde{\rho}$.\\
	\begin{algorithmic}[1]
		\STATE $\rho_2^0\gets 0 $;
		\STATE $k \gets 0$;
		\WHILE{$k \leq n_{\mathrm{it}}$}
		\STATE $\rho_1^{k+1}\gets$ ConjGrad$\left (C_h^{T} C_h + \nu_k\mathrm{Id}\, , C_h^{T} u + \nu_k \rho_2^k \right )$;
		\STATE $\sigma_{k+1} \gets $ Noise-Estimator$(\rho_1^{x+1})$;
		\IF {$k=0$}
		\STATE $\lambda \gets\nu_0 \cdot\sigma_0^2$
		\ENDIF
		\STATE $\rho_2^{k+1}\gets$Denoise$\left (\tilde{\rho}_1^{k+1}\, ,\sigma_{k+1}\right ) $; \hfill\COMMENT{ZeroShot-Denoiser}
		\STATE $\nu_{k+1}\gets \lambda /\sigma_{k+1}^2$;
		\STATE $k\gets k+1$;
		\ENDWHILE
		\RETURN $\tilde{\rho}_2^{k+1}$
	\end{algorithmic}
\end{algorithm}

\section{Experimental Results}\label{sec:experiments}

In this section of the paper we describe the experimental results obtained on real MPI FPP data. In particular, we show the results obtained on two different datasets, one with a standard Lissajous-type scanning trajectory and one with a non-Lissajous trajectory. In Experiment 1 (\ref{subsec:data}) we reconstruct 6 different phantoms and compare the results obtained with state-of-the-art methods using both measured and model-based approaches. The dataset used is described in \ref{subsec:data} and the preprocessing employed for each channel. In \ref{subsec:percentile} we describe a percentile-based idea that helps reducing artifact in the deconvolution. To showcase the flexibility of our method and its path-independence, we show in \ref{subsec:exp2} the reconstructions obtained with custom data obtained employing sinusoidal dynamic fields with the superposition of a fast-oscillating excitation field. Moreover, the reconstructions are performed only on the signal collected on the $x$-axis, showing the robustness of our method to partial data as well as to different scanning trajectories. All experiments in this paper where run on a workstation with 13th Gen Intel(R) Core(TM) i9-13900KS, 128 GB of Ram, and NVIDIA RTX A6000 GPU and Windows 11 Pro.

\subsection{Data}\label{subsec:data}

We consider the 2D FFP ``MPIdata: Equilibrium Model with Anisotropy" dataset (EMWA) \cite{knopp2024equilibriumdata} published with \cite{maass2024equilibriumanysotropy}. In this datasets, 6 different phantoms have been scanned the preclinical MPI scanner from Bruker (Ettlingen, Germany). The scanner employs co-sinusoidal scanning trajectories of the form
\begin{equation}\label{eq:lissajous}
	r(t) = (A_x\cos(2\pi f_x t))\, , A_y\cos(2\pi f_y t)\, , 0)^T .
\end{equation}

The gradient is $G = \diag (-1,-1,2)\si{\tesla\per {(\meter \mu_0})}$ and the drive frequencies are set to $f_x = \frac{2.5}{102}\si{\mega\hertz} $ and $f_y = \frac{2.5}{96}\si{\mega\hertz}$ with drive field amplitudes of nominal value $A_x = A_y = 12\si{\milli\tesla}/\mu_0$. This corresponds to a scanning area (the so called Drive-Field Field of View DF-FoV) of $24\si{\milli\meter}\times 24\si{\milli\meter}$.

The used phantom are displayed in Fig.~\ref{fig:exp1}. The snake is composed of five cubic rods ($2.5\si{mm}\times 2.5\si{mm}$ cross-section) with lengths of $20\si{mm}$, $17.5\si{mm}$, $15\si{mm}$, $8.75\si{mm}$ and $5\si{mm}$, arranged in a winding, snake-like pattern. The ice-cream phantom features a lower cone topped with a  spherical upper part. The dot phantom is the delta-sample, positioned $6\si{mm}$ away from the center along both the $x$ and $y$-axes. The three resolution phantoms resemble an equality sign and they are formed by two rods of lengths $20\si{mm}$ and $17.5\si{mm}$ positioned at $7\si{mm}$ (Res. 3), $5\si{mm}$ (Res. 2) and $3\si{mm}$ (Res. 1), respectively. All phantoms have been filled with the tracer perimag (micromod, Rostock, Germany) with a concentration of $10\si{\milli\gram}_{\mathrm{Fe}}\si{\milli\per\liter}$. The provided system matrix has been collected by moving the delta-sample on a $17\times 15$ grid on a $34\si{\milli\meter}\times 30\si{\milli\meter}$ FoV, which is bigger than the DF-FoV and contains calibration of voxels outside the scanning area. An overview of the parameters used in the experiment can be found in Tab. \ref{tab:params_si}.
\begin{table}[h]
	\centering
	\small
	\begin{tabular}{ll}
		\hline
		\textbf{Parameter} & \textbf{Value} \\
		\hline
		\multicolumn{2}{l}{\textbf{Constants}} \\
		Magnetic permeability $\mu_0$ & $4\pi \times 10^{-7} \si{\henry\per\meter}$ \\
		Boltzmann constant $k_b$ & $1.38064852 \times 10^{-23} \si{\joule\per\kelvin}$ \\
		
		\multicolumn{2}{l}{\textbf{Scanner}} \\
		Base frequency & $2.5 \si{\mega\hertz}$ \\
		Frequency dividers & $(102,\, 96,\, 99 )$\\
		Amplitudes $A_x = A_y $& $0.012 \si{\tesla}/\mu_0$ \\
		Repetition time & $ 6.528 \times 10^{-4} \si{\second}$ \\
		Sampling points &  1632\\
		Gradient strength $\mu_0 G$ & $\diag (-1,\, -1,\, 2) \si{\tesla\per\meter}$ \\
		
		\multicolumn{2}{l}{\textbf{Particle}} \\
		Temperature $T_B$ & $293 \si{\kelvin}$ \\
		Saturation magnetization $\msat$ & $4.74 \times 10^{5} \si{\joule\per\cubic\meter\tesla}$ \\
		Core diameter $d$ (average) & $21  \si{\nano\meter}$ \\
		\hline
	\end{tabular}
	\caption{\small Model parameters used in the reconstruction.}
	\label{tab:params_si}
\end{table}

\subsection{Preprocessing}\label{subsec:preprocess}

As a first preprocessing step, we divide the Fourier transform $\hat{s}$ of the signal by the transfer function $\hat{a}$ as done in \cite{KnoppBiederer_etal2010}. For the computation of the transfer function, we consider the system matrix provided with the EMWA dataset; following \cite{KnoppBiederer_etal2010}, we compute the transfer function in a least square fashion using in-house simulations for comparison. 
Furthermore, using the same pairs of measured and simulated system matrices we compute Signal-to-Noise ratios (SNR) for each frequency; this SNR is used to perform SNR thresholding of the data, as some frequencies are unreliable. This is a standard procedure (cf. \cite{maass2024equilibriumanysotropy,sanders2025physicsbased}) as it helps with signal attenuation and recovers phase shifts which are introduced by amplifiers and filters \cite{vonGladiss2020}.

\begin{figure}[t]
	\centering
	\includegraphics[width=\columnwidth]{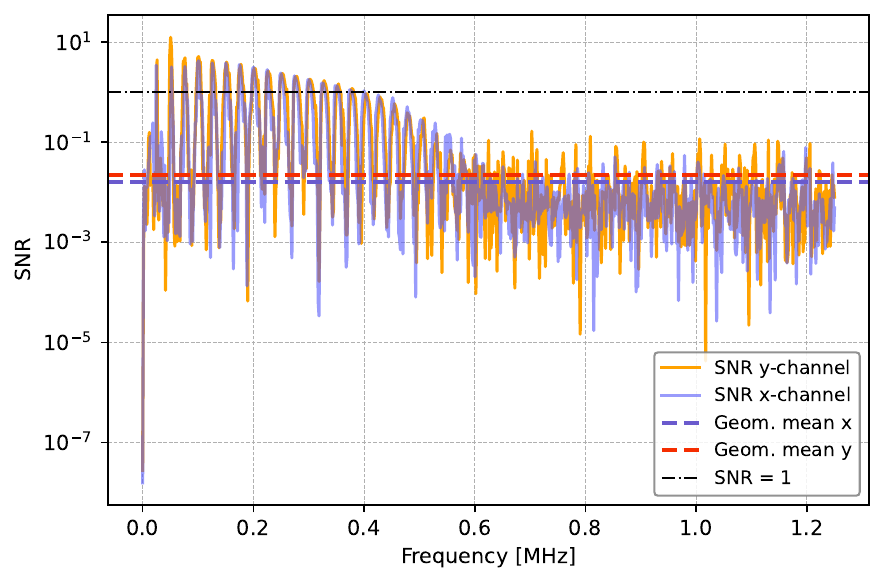}
	\caption{\small Logarithmic plot of the SNR across frequencies for the $x$ and $y$ channels. The $y$-channel exhibits higher SNR (Wilcoxon signed-rank test, $p\approx 10^{-9}<0.05$). Geometric means are shown as dotted lines. The dashed gray line at SNR = 1 indicates the threshold where the signal equals noise in power level.}
	\label{fig:snr}
\end{figure}

To obtain good quality results, the SNR threshold has been chosen to different levels $\Theta_x$ and $\Theta_y$ for each channel. This is justified by the observation that the frequencies on the $y$-channel are more reliable than the ones on the $x$-channels; this is shown in Fig. \ref{fig:snr}, where we plot both the SNR values, and corroborated by the low $p$-value ($\sim10^{-9}$) upon performance of a paired statistical test (Wilcoxon signer rank test \cite{wilcoxon1945test}) across all frequency components (n = 817). Moreover, they have to be chosen \emph{ad hoc} for each phantom: while for the resolution 1,2 and 3 phantoms there was no thresholding necessary ($\Theta_x =\Theta_y=0$), for every other phantom $\Theta_x = 0.04$ and $\Theta_y = 0.01$. 

\subsection{Denoiser Artifact Reduction}\label{subsec:percentile}

\begin{figure}[t]
	\centering
	\includegraphics[width=\columnwidth]{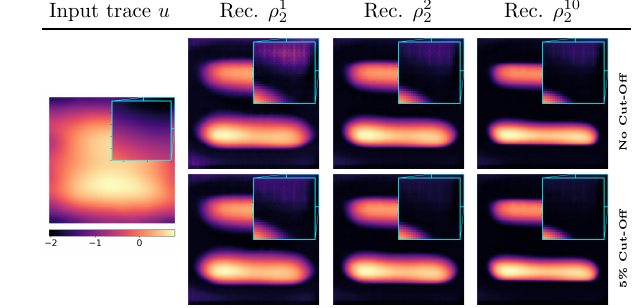}
	\caption{ Shared trace image (left) and Plug-and-Play reconstruction outputs after $k=0$, $k=1$, and $k=9$ iterations for the Res. 2 phantom (cf. Fig. \ref{fig:exp1}) with our algorithm. Labels on the right indicate the use of no cut-off (top) and lower 5\% percentile cut-off (bottom). The percentile strategy improves artifact suppression.} 	
	\label{fig:cutoff}
\end{figure}

In this section we describe a percentile-based strategy of artifact reduction in the Deconvolution Stage. Indeed, in model-based MPI applications, it is possible to have regions of the image in which the values of the reconstructions get pushed into relatively deep negative values (cf. the trace $u$ in Fig. \ref{fig:cutoff}). On the other hand, the denoiser employed in the ZeroShot-PnP is trained on a large dataset of natural images with non-negative pixel values \cite{Zhang2022pnp}. In our experiments we have observed that these negative-valued regions might create artifacts in the reconstruction (see for example the zoom-ins in the first row of Fig. \ref{fig:cutoff}). To mitigate this effect, we perform percentile-based trimming on the variables $\rho_1^{k+1}$ at each iteration: we clip the concentration range of $\rho_1^{k+1}$ by discarding the values below the $5^{th}$ percentile. In Fig. \ref{fig:cutoff} we show the difference between no trimming (first row) and $5\%$ percentile trimming (second row) for the first, second, and last iterations. We observe that already after the first iteration, the artifact has been reduced and completely suppressed by the time the method hits the last iteration.

\subsection{Experiment 1: Comparison with State-of-the-art Methods}\label{subsec:exp1}

In this section we show the reconstructions obtained with the algorithm presented in this paper on the EMWA dataset described in \ref{subsec:data}.

The comparison methods considered are:
\begin{itemize}
	\item[i)]\textbf{Meas. SM:} is the Tikhonov inversion of the system matrix provided with the EMWA dataset and obtained via calibration of the FoV on a $17\times 15$ grid. The inversion is performed with the Kazcmarz method with the parameters described in \cite{droigk2022multidimcheb}.
	\item[ii)] \textbf{EQ-SM:} the reconstruction is performed by inversion of a system matrix whose ($17\times 15$) columns are simulated using the Langevin model in the FoV. The inversion is performed with the Kazcmarz method \cite{droigk2022multidimcheb}.
	\item[iii)] \textbf{DCR-EQ:} the reconstruction is performed using the direct Chebychev reconstruction method presented in \cite{droigk2025efficientcheb,droigk2022multidimcheb}. The reconstruction is performed on a $21\times 21$ grid in the DF-FoV following \cite{droigk2025efficientcheb} and uses the Langevin model.
\end{itemize}
\begin{figure}[t!]
	\centering
	\includegraphics[width=\columnwidth]{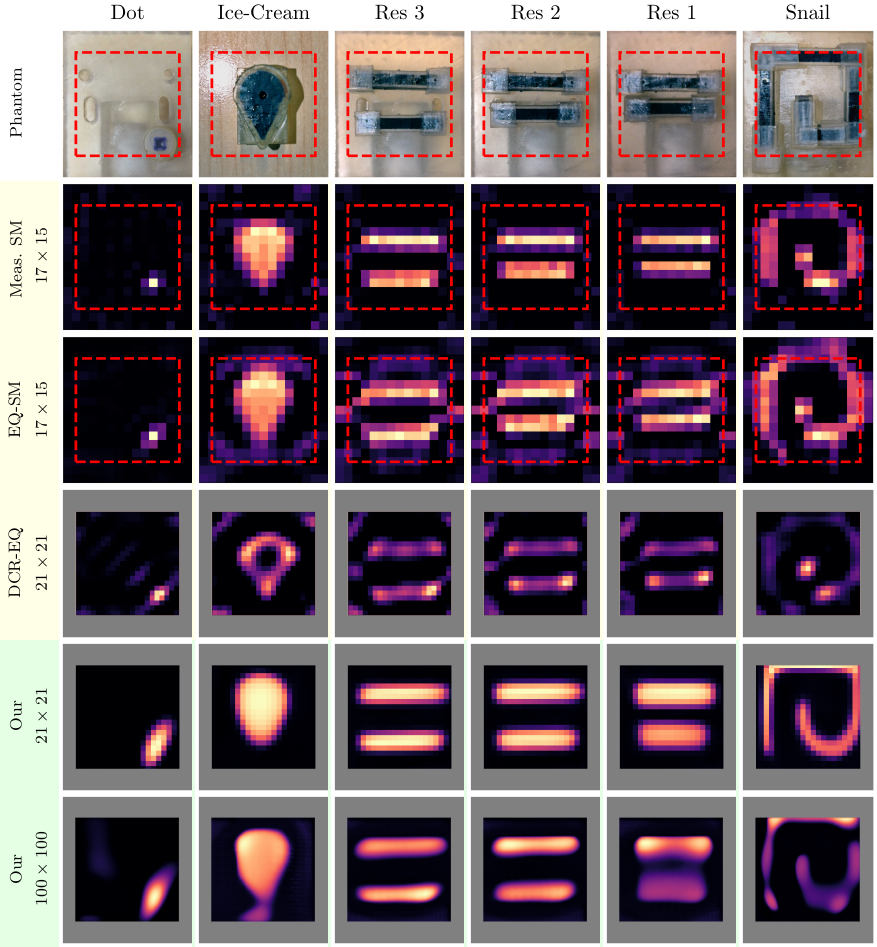}
	\caption{\small Reconstruction results for the phantoms (Dot, Ice-Cream, Resolution targets, Snail) in the EMWA dataset \cite{knopp2024equilibriumdata,maass2024equilibriumanysotropy}. Each row shows reconstructions obtained with a different reconstruction approach: inversion of the measured system matrix (Meas. SM), inversion of a simulated system matrix (EQ-SM), the direct reconstruction with Chebyshev polynomials \cite{droigk2025efficientcheb} (DCR-EQ) and the methods presented in this paper (Our). The red bounding box indicates the DF-FoV, both in the ground truth images and the reconstructions obtained using the system matrix over the full FoV. Conversely, reconstruction computed within the DF-FoV are padded with a gray area to match the full FoV dimensions. The system-matrix uses a $17\times 15$ resolution on the FoV, stemming from the size of the available calibration data. The DCR-EQ method is reconstructed as in \cite{droigk2025efficientcheb}  on a $21\times 21$ grid on the DF-FoV. The results with our method are provided both on a $21\times 21$ grid for comparison (second to last row) and on an increased $100\times 100$ grid size (last row), in both cases on the DF-FoV. The pictures of the phantom in the first row are available with the EMWA dataset and are courtesy of \cite{maass2024equilibriumanysotropy,knopp2024equilibriumdata}.
	}
	\label{fig:exp1}
\end{figure}
We have performed reconstructions with the baseline methods described, and with our method, for which we show reconstructions in the DF-FoV on both a $21\times 21$ grid and on a $100\times 100$ grid. An overview of all the reconstruction is shown in Fig. \ref{fig:exp1}. We remark that the DF-FoV is outlined by the scanning trajectory and contains the scanning points, whereas the FoV contains a boundary area around the DF-FoV without data points. Consequently, the ground truth images and the methods employing the FoV contain the bounding box representing the DF-FoV (in red in Fig. \ref{fig:exp1}), whereas the methods using the DF-FoV have been padded with gray values up to the FoV boundary to preserve relative size of all reconstructions shown. Concerning preprocessing and parameters used in the comparison methods, we adopt the preprocessing steps and parameter settings as recommended in their respective publications \cite{maass2024equilibriumanysotropy,knopp2024equilibriumdata,droigk2025efficientcheb}. We observe that, compared with the other methods using the Langevin model (EQ-SM, DCR-EQ), our algorithm is able to reconstruct the underlying particle distribution and with rather clean images, especially on the $21\times 21$ grid. Even if the scanning trajectory in the EMWA is rather sparse (cf. Fig. \ref{subfig:traj:liss}), we have reconstructed the phantoms on a $100\times 100$ grid, to showcase the inpainting effect of the variational Core Stage in \eqref{eq:first:step}.
The reconstructions wall-clock times on the CPU and GPU for both $21\times 21$ and $100\times 100$ reconstructions grid are collected in Tab. \ref{tab:runtimes}.

\begin{table}[ht]
	\centering
	\begin{tabular}{>{\centering\arraybackslash}m{1.5cm} c c}
		%\toprule
		Grid-size& \textbf{CPU} & \textbf{GPU} \\
		\midrule
		$21\times 21$ & $0.46\si{\second}$ & $1.56\si{\second}$ \\
		$100\times 100$ & $5.84\si{\second}$ & $3.36\si{\second}$ \\
		%\bottomrule
	\end{tabular}
	\caption{Runtime comparison between CPU and GPU for our algorithm on the EMWA dataset.}
	\label{tab:runtimes}
\end{table}

\subsection{Experiment 2: Reconstruction with non-Lissajous Trajectories}\label{subsec:exp2}

\begin{figure}[t]
	\centering
	\begin{subfigure}[t]{0.49\columnwidth}
		\includegraphics[width=\linewidth]{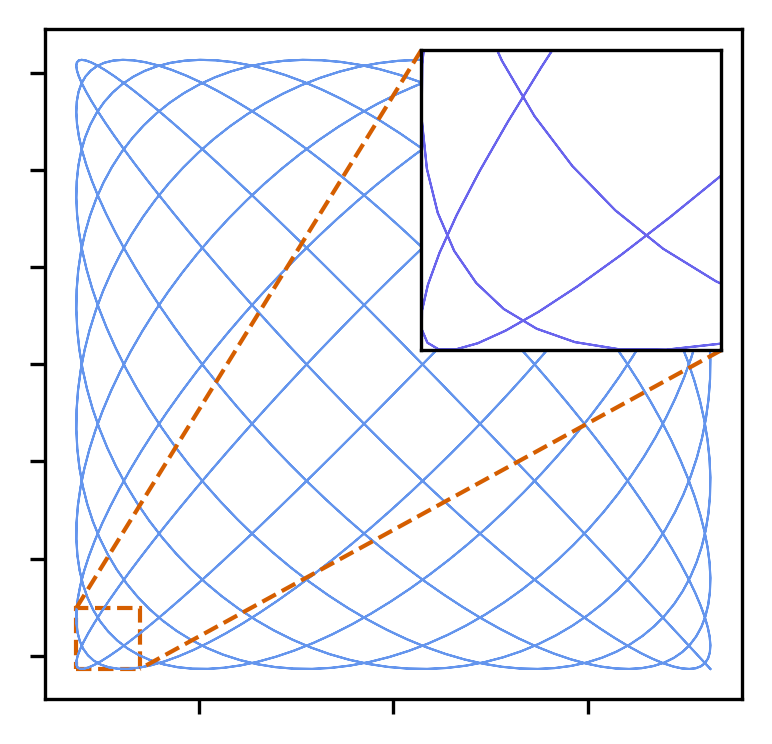}
		\caption{Lissajou trajectory}
		\label{subfig:traj:liss}
	\end{subfigure}
	\begin{subfigure}[t]{0.49\columnwidth}
		\includegraphics[width=\linewidth]{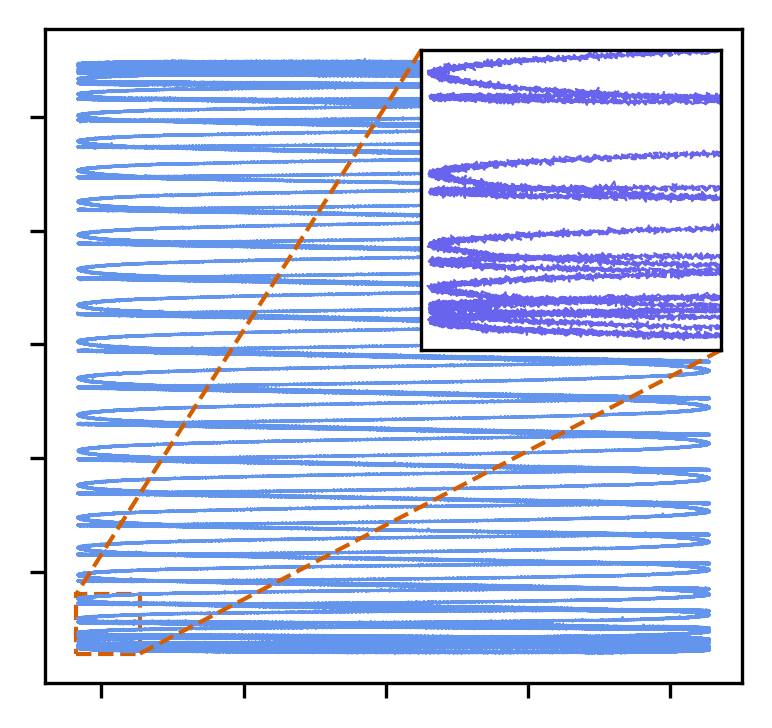}
		\caption{Non Lissajous trajectory}
		\label{subfig:traj}
	\end{subfigure}
	\caption{\small Comparison between the Lissajous trajectory \ref{subfig:traj:liss} in the EMWA Dataset and the non-Lissajous trajectory \ref{subfig:traj} obtained adding an excitation field in experiment 2.}
	\label{fig:traj}
\end{figure}

In this section we show the reconstruction from data obtained with the scanner described in \cite{leili2025transverseMNP}. This scanner employs a 2D FFP scan with a gradient of approximately $1.39\si{\tesla\per\meter}$ on the $x$-axis and $-3.16\si{\tesla\per\meter}$ along the $y$-direction. The frequencies for the drive field are $f_x = 50\si{\hertz}$ and $f_y = 1\si{\hertz}$ with a FoV of approximately $32\si{\milli\meter}\times 30\si{\milli\meter}$. Additionally and differently from the data in experiment 1, a high-frequency 1D excitation field has been added along the $x$-direction with a frequency $f_e = 25\si{\kilo\hertz}$. Consequently, the scanning trajectory is not of the Lissajous type, as it moves along a superposition of sinusoidal curves along the $x$-axis
\begin{equation}
	r(t) = \begin{pmatrix}
		A_x \sin(2\pi f_x t) + A_e \sin(2\pi f_e t ) \\
		A_y \sin(2\pi f_x t)
	\end{pmatrix},
\end{equation}
for amplitudes $A_x$, $A_y$ and $A_e$ (we have not modeled the amplitudes as we have directly the current data at our disposal).
The sample frequency is set to $6\si{\mega\hertz}$, resulting in a great quantity of data points. The plot of the trajectory as well as a comparison with the Lissajous trajectory in experiment 1 is shown in Fig. \ref{fig:traj}. In this experiment, instead of simulating the scanning trajectory, we have employed the current data provided with the scans to obtain the realistic scanning trajectory in Fig. \ref{subfig:traj}.
The velocities have been computed by forward differences on the data points. As an additional difference, we only have the signal collected from receive coils along the $x$-axis. This means that, even though the velocities of $r(t)$ span $\mathbb{R}^2$ (cf. Fig. \ref{subfig:traj}) from a mathematical standpoint, the second row is not accessible and the trace of the Core Operator cannot be computed. Indeed, the relationship in \eqref{eq:core:formula} reduces to an equation of the form
\begin{equation}
	s_x (t) =
	\begin{pmatrix}
		A_{0,0} & A_{0,1}
	\end{pmatrix} 
	\begin{pmatrix}
		v_x (t) \\ 
		v_y (t)
	\end{pmatrix}.
\end{equation}
The lack of data $s_y$ along the $y$-channel offers no obstruction, as we simply reconstruct the first diagonal entry $A_{0,0}$ of the MPI Core Operator in the Core Stage, and deconvolve it using the first diagonal entry of the MPI kernel matrix in \eqref{eq:kernel:matrix} \cite{marz2022icnaam}. 

The phantoms employed in this study consist of two resolution phantoms akin to those in experiment 1, filled with Synomag-D with particle diameter of $70\si{\nano\meter}$. The two bars are put at a distance of $3\si{\milli\meter}$ , $2\si{\milli\meter}$ and $1\si{\milli\meter}$, respectively.
\begin{figure}[t]
	\centering
	\includegraphics[width=\columnwidth]{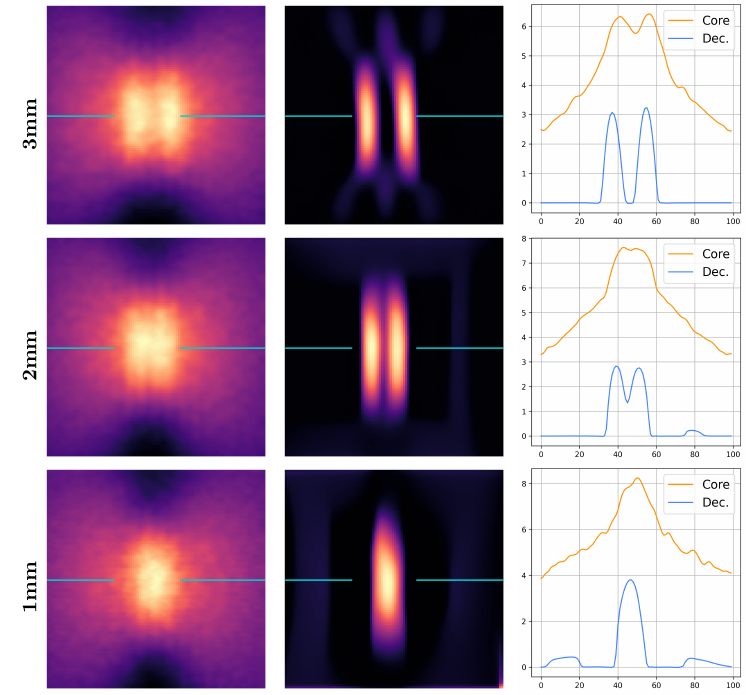}
	\caption{\small From left to right: output of the Core Stage $(A_h )_{0,0}$, its deconvolution with the ZeroShot-PnP algorithm (central column) and the extracted profiles (right column). Horizontal profiles (in cyan) through the reconstructed images at the center slice for the $3\si{\milli\meter}$ (top row), $2\si{\milli\meter}$ (middle row) and $1\si{\milli\meter}$. }
	\label{fig:exp2}
\end{figure}
The results can be seen in Fig. \ref{fig:exp2}, where we have also plotted a 1D slice of the Core Stage output and of its deconvolved version. We can see that with our method, it is possible to distinguish between the bars up to a distance as small as $2\si{\milli\meter}$ using a fully model-based reconstruction. We remark that in this experiment no preprocessing of the type described in \ref{subsec:preprocess}: no SNR-thresholding and no transfer function. Moreover, we remark that, due to the high number of datapoints obtained with a $6\si{\mega\hertz}$ sample frequency, we have only used $1\%$ (one point every 100) of the data available to increase the speed of the reconstruction and the convergence of the methods. For these phantoms we have kept the reconstruction parameters of Experiment 1 for the Core Stage, whereas we have run a grid search for both $\hsat$ and $\nu_0$ for the deconvolution stage and selected $\hsat = 10^{-2}/\mu_0$ and $\nu_0 = 4\cdot 10^{-2}$.

\section{Conclusion}

In this paper we have presented the first reconstruction results in FFP MPI obtained with a model-based approach that is independent from the chosen scanning trajectory. In a first experiment, we have shown that the proposed method is robust and yields good quality results by using it on the publicly available 2D Equilibrium-Model with Anisotropy dataset \cite{knopp2024equilibriumdata}. The quality of the reconstructions is highlighted by the comparison with state-of-the-art baseline methods described in \ref{subsec:exp1}. As a second experiment, we have shown the flexibility of the method by showing the reconstruction obtained with the scanner in \cite{leili2025transverseMNP}. This scanner differs from the Bruker's scanner as it employs an additional high-frequency oscillating field and that it collects data only along the $x$-channel of the scanner. We have provided a study on three different resolution phantoms.

We point out that model-based reconstruction based on the MPI Core Operator has been shown to be beneficial in various simulated scenarios. Such scenarios include reconstruction with merged trajectories for quality-enhancing purposes \cite{gapyak2022mdpi} as well as in multi-patching scenarios \cite{gapyak2023multipatch}. Recent results have also extended the model-based reconstruction formulae to 3D Field-Free Line MPI \cite{gapyak2023ffl3d}. Therefore, future directions of research include the adoption and evaluation of the corresponding methods in such scenarios.
Additionally, in this work we have used the Langevin model for paramagnetism which offers a relatively simple description of the phenomena involved. Real data are however affected by noise and show relaxation effects. Advanced approaches try to model such relaxation effects by incorporating Brownian and N\'eel rotations in the computation of magnetic moments by solving the Fokker-Plank equations \cite{kluth2018mathematical}. Combination of more complex models as in \cite{maass2024equilibriumanysotropy} with the methods here presented are a direction of future investigation.

\section*{Acknowledgment}

This work was supported by the Hessian Ministry of Higher Education, Research, Science and the Arts within the Framework of the ``Programm zum Aufbau eines akademischen Mittelbaus an hessischen Hochschulen" and by the German Science Fonds DFG under grant INST 168/4-1.
The authors would like to thank Prof. Guang Jia and Prof. Shouping Zhu, along with their research teams at Xidian University, Xi'an, Shaanxi, China, for providing access to the data used in Experiment 2 of this work and for insightful discussions.

\bibliographystyle{ieeetr}
\bibliography{literature}

\end{document}